# Automated Artifact Detection in Ultra-widefield Fundus Photography of Patients with Sickle Cell Disease

Anqi Feng, MSc[1,*]; Dimitri Johnson, BS[2,3,*]; Grace R. Reilly, MD[2,4]; Loka Thangamathesvaran, MD[2]; Ann Nampomba, MSc[2]; Mathias Unberath, PhD[2,5,6]; Adrienne W. Scott, MD[2]; Craig Jones, PhD[2,5,6]

[1]Department of Biomedical Engineering, Johns Hopkins School of Medicine, Baltimore MD

[2]Wilmer Eye Institute, Johns Hopkins School of Medicine, Baltimore MD

[3] Morehouse School of Medicine, Atlanta GA

[4] Drexel University College of Medicine, Philadelphia PA

[5]Department of Computer Science, Johns Hopkins School of Medicine, Baltimore MD

[6]Malone Center for Engineering in Healthcare, Johns Hopkins School of Medicine, Baltimore MD

* Co-first authors

Corresponding Authors:

Craig Jones, PhD
Malone Hall, Suite 340
3400 North Charles Street
Baltimore, MD 21218-2608, USA
(410) 449-2024
craigj@jhu.edu

Adrienne W. Scott, MD
Retina Division, Wilmer Eye Institute,
Johns Hopkins University School of Medicine
600 North Wolfe Street, Maumenee 719
Baltimore, MD 21287, USA
410-502-9247
ascott28@jhmi.edu

## Key Points

**Question:** Can deep learning automate detection of artifacts in ultra-widefield color fundus photographs of sickle cell patients?

**Findings:** In a cross-sectional study using 243 ultra-widefield color fundus photographs from patients with proliferative sickle cell retinopathy, a convolutional neural network was trained to detect image artifacts. Accuracy for each class was Eyelash Present at 83.7%, Lower Eyelid Obstructing at 83.7%, Upper Eyelid Obstructing at 98.0%, Image Too Dark at 77.6%, Dark Artifact at 93.9%, and Image Not Centered at 91.8%.

**Conclusions and Relevance:** With further refinement, automated detection of ultra-widefield color fundus photograph artifacts will enable immediate detection of artifacts appropriate to minimize return visits in remote settings and applicable in large trials for an objective, efficient artifact detection.

# Abstract

**Importance:** Ultra-widefield fundus photography (UWF-FP) has shown utility in sickle cell retinopathy screening; however, image artifact may diminish quality and gradeability of images.

**Objective:** To create an automated algorithm for UWF-FP artifact classification.

**Design:** A neural network based automated artifact detection algorithm was designed to identify commonly encountered UWF-FP artifacts in a cross section of patient UWF-FP. A pre-trained ResNet-50 neural network was trained on a subset of the images and the classification accuracy, sensitivity, and specificity were quantified on the hold out test set.

**Setting:** The study is based on patients from a tertiary care hospital site.

**Participants:** There were 243 UWF-FP acquired from patients with sickle cell disease (SCD), and artifact labelling in the following categories was performed: Eyelash Present, Lower Eyelid Obstructing, Upper Eyelid Obstructing, Image Too Dark, Dark Artifact, and Image Not Centered.

**Results:** Overall, the accuracy for each class was Eyelash Present at 83.7%, Lower Eyelid Obstructing at 83.7%, Upper Eyelid Obstructing at 98.0%, Image Too Dark at 77.6%, Dark Artifact at 93.9%, and Image Not Centered at 91.8%.

**Conclusions and Relevance:** This automated algorithm shows promise in identifying common imaging artifacts on a subset of Optos UWF-FP in SCD patients. Further refinement is ongoing with the goal of improving efficiency of tele-retinal screening in sickle cell retinopathy (SCR) by providing a photographer real-time feedback as to the types of artifacts present, and the need for image re-acquisition. This algorithm also may have potential future applicability in other retinal diseases by improving quality and efficiency of image acquisition of UWF-FP.

# Introduction

Sickle cell disease (SCD) comprises a group of hemoglobinopathies that comprise most common inherited blood disorder, affecting over 100,000 in the United States and more than 300,000 individuals globally[1]. Abnormal hemoglobin leads to impaired oxygen transport, causing chronic systemic vaso-occlusion and endothelial dysfunction. The most common cause of vision impairment and blindness in SCD occurs from proliferative sickle cell retinopathy (PSR) when pathologic sea fan neovascularization develops because of retinal ischemia. While untreated sea fan neovascularization may auto-infarct spontaneously without visual consequence in upwards of 30% of eyes with PSR,[2] PSR can progress to vitreous hemorrhage and/or retinal detachment which may require surgical intervention, from which poor visual outcomes have been described[3]. Though the literature is limited with regard to PSR treatment, a randomized, prospective controlled trial reported decreased rates of vision loss and PSR progression in eyes treated with scatter laser photocoagulation compared to control eyes.[4] Therefore, it is important to develop and refine PSR screening methods to identify potentially-sight threatening sea-fan neovascularization early while the patient remains asymptomatic, as treatment may prevent vision loss.

Current expert consensus guidelines for sickle cell retinopathy suggest screening patients with SCD with dilated fundus exam by age 10, with 1-2 year follow up thereafter depending upon clinical findings.[5] Adults with SCD and caregivers of children with SCD report missing appointments,[6] and adherence to these retinopathy screening guidelines may be as low as 40%.[7] Therefore, multi-disciplinary appointments with point of care testing may improve adherence to screening recommendations.[8]

Ultra-widefield fundus photography (UWF-FP), in which greater than 200 degrees of the retinal periphery may be captured in a single image, has shown utility in retinopathy screening in undilated SCD patients in a U.S. Hematology clinic, with retinal images obtained by novice photographer clinic personnel.[9]

The utility of ultra-widefield fundus photography has also been reported as effective in facilitating retinopathy screening through tele-medicine approaches in a variety of retinal conditions, including proliferative diabetic retinopathy (PDR), retinopathy of prematurity (ROP), uveitis, and retinal vasculitis.[10] The utility of UWF-FP in screening for retinal disease is dependent upon image quality and may be diminished by image artifacts. For example, the large depth of focus provided by the Optos camera can often result in eyelash or nose artifacts, as well as peripheral distortion of images.[10]

The prevalence of SCD is highest in medically underserved areas. For example, the annual disease burden of SCD in sub-Saharan Africa is more than 80 times of that noted in the United States.[11] Therefore, UWF-FP screening is also a particularly attractive global approach for SCR detection for geographic areas that are likely to have retinopathy screening performed by novice photographers. Because SCR primarily affects the peripheral retina, minimizing artifacts, particularly in the distal regions of the UWF-FP, is critical to optimize image capture of the areas of pertinent pathology.

Artifact detection of PSR imaging using machine learning algorithms is a novel area of exploration. Herein, we describe a novel method for automated artifact detection in a cohort of SCD patients using UWF-FP: 1) applies artifact labels to UWF-FP images automatically, 2) enables accurate artifact determination useful for fast artifact determination appropriate for imaging acquired in large urban centers or remote locations, and 3) the algorithm will give the photographer immediate feedback for repeat scans to minimize the need for the patient to come for a return visit.

## Methods

### Data Acquisition

The dataset used in this study was obtained from consecutive patients with SCD enrolled from a hematology clinic at Johns Hopkins Hospital between April 2018 and March 2021. Images were taken without pupillary dilation using an Optos UWF Primary camera (Optos, Inc.).[9] The dataset consisted of 243 images, which were divided into three sets with a 7:1:2 ratio for training, validation, and testing, respectively. Prior to training, all images were normalized within the range of [-1,1] for three color channels. The images were then resized from their original dimensions of 4000 × 4000 × 3 to 224 × 224 × 3. Additionally, to increase the variability of the training data and to reduce overfitting, we applied random affine transformations and random horizontal flipping to the training set.

### Image Grading

The dataset images (N=243) were graded by an expert retina specialist grader (A.W.S.) for the presence of the following six commonly observed UWF-FP artifacts: eyelash present (N=232), lower eyelid obstructing (N=41), upper eyelid obstructing (N=200), image too dark (N=83), dark artifact (N=204), and image not centered (N=99). The image grader was masked to all other patient information. Another study team member presented de-identified images to the grader and collated artifact labelling. This expert grading was used as "ground truth" to which the algorithm performance was compared.

Network Structure

In this study, we adopted the ResNet-50[12] to perform the multi-label classification task. The network structure is shown in Figure 1. ResNet-50 is a common convolutional neural network that has achieved state-of-the-art performance on multiple image classifications tasks. ResNet-50 contains 50 layers which are organized into several stages, each stage contains multiple residual blocks, which consist of convolutional layers, batch normalization layers[13], ReLU activation functions[14], and shortcut connections. The residual blocks help to mitigate the problem of vanishing gradients and enable the training of deeper neural networks. The network will take in a batch of retina images, where each image has a shape of 224 × 224 × 3 and generate a vector of six probabilities representing the likelihood of the input image belonging to each of the six classes.

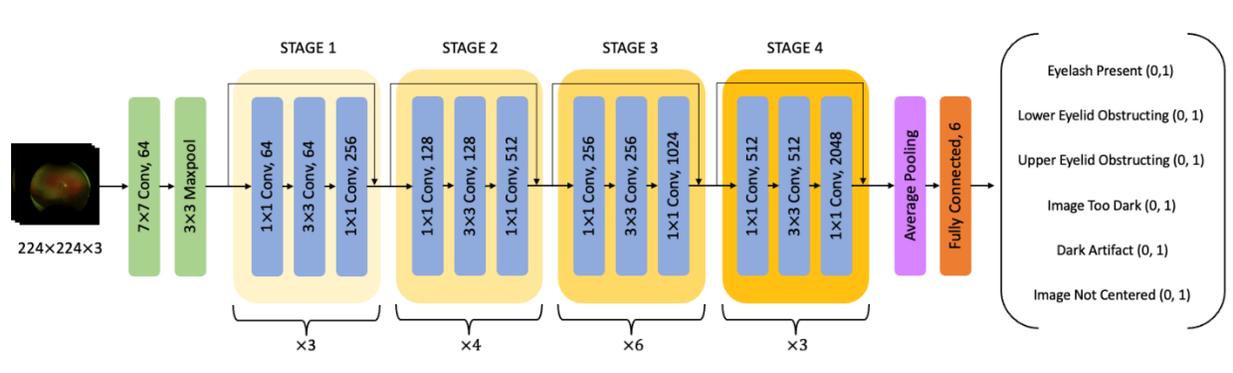

Figure 1: Resnet-50 neural network structure for the labeling of the six classes of artifacts noted for this work.

To enhance the performance of our model, we utilized transfer learning based on a pre-trained ResNet-50 model that was trained on the ImageNet dataset[15] to initialize the network parameters. By leveraging the knowledge gained from the large and diverse ImageNet dataset, the pre-trained model provided a strong starting point for our specific classification task.

Implementation Details

In our study, we initially employed a 5-fold cross-validation approach. By analyzing the consistent performance across all folds, we identified the most suitable set of hyperparameters. The loss function was a weighted binary cross entropy loss to account for the class imbalance. We used the SGD optimizer to minimize the loss with an initial learning rate of $5 \times 10^{-3}$, which decreased by 10% every 10 epochs, and a momentum of 0.9. After determining the hyperparameters, we trained the neural network on the entire dataset, excluding the test data, to further enhance its performance and generalization capabilities. The model was scheduled to be trained for 500 epochs but used early stopping (patience = 5). In the evaluation stage, we computed the average accuracy, precision, recall, specificity, and sensitivity for all six classes. These metrics were used to assess the performance of our model and determine its effectiveness in correctly identifying each class.

All code was written in PyTorch and run on GPU cluster that contained two P100 NVIDIA GPUs.

# Results

The preliminary 5-fold validation results are shown in Table 1. The results indicate that the model demonstrates relatively consistent performance across all folds under a set of optimal hypermeters.

Table 1: Accuracy (in percent) of the hold out set over the 5-fold cross-validations for the optimal hyper-parameters.

|  | Eyelash Present | Lower Eyelid Obstructing | Upper Eyelid Obstructing | Image Too Dark | Dark Artifact | Image Not Centered |
|---|---|---|---|---|---|---|
| Fold 1 | 83.7 | 81.6 | 95.9 | 75.5 | 85.7 | 93.9 |
| Fold 2 | 81.6 | 79.6 | 93.9 | 75.5 | 87.8 | 93.9 |
| Fold 3 | 85.7 | 83.7 | 98.0 | 73.5 | 85.7 | 93.9 |
| Fold 4 | 85.7 | 83.7 | 93.9 | 71.4 | 87.8 | 93.9 |
| Fold 5 | 87.8 | 79.6 | 93.9 | 71.4 | 87.8 | 95.9 |
| **Overall** | **84.9+/-2.1** | **81.6+/-1.8** | **95.1+/-1.6** | **73.5 +/- 1.8** | **87.0 +/- 1.0** | **94.3 +/- 0.80** |

We then assessed the performance of our classification model on the test set to determine its accuracy. The average accuracy scores for each class were as follows: Eyelash Present at 83.7%, Lower Eyelid Obstructing at 83.7%, Upper Eyelid Obstructing at 98.0%, Image Too Dark at 77.6%, Dark Artifact at 93.9%, and Image Not Centered at 91.8%. These results demonstrate that our model exhibits a high degree of proficiency in identifying the presence of upper eyelid obstructions, dark artifacts, and off-centered images, while showing room for improvement in detecting cases with lower eyelid obstructions, eyelashes, and excessively dark images.

In addition to accuracy, we computed several other evaluation metrics to provide a more comprehensive understanding of our model's performance. The average precision was calculated to be 87.1%, indicating a strong ability to correctly identify relevant instances among the retrieved results. The average recall and sensitivity, both measuring the proportion of true positive cases identified, were found to be 84.2%. This suggests that our model effectively recognizes the majority of the cases exhibiting the target features. Furthermore, the average specificity of 92.7% demonstrates the model's robust performance in correctly identifying true negative cases and distinguishing them from false positives. These evaluation metrics collectively highlight the overall reliability and discriminative power of our model in analyzing ocular images.

We then evaluated the performance of the model using the confusion matrix, which provided a detailed assessment of prediction accuracy across various class labels. For instance, as illustrated in Figure 2, in the "Eyelash Present" class, the model recorded 26 true positives, 14 true negatives, 4 false negatives, and 5 false positives, indicating effective recognition of eyelash presence, but with room for improvement in reducing misclassifications. In the "Lower Eyelid Obstructing" class, despite a relatively high true negative rate, the model struggled to detect cases where the lower eyelid obstructed the image, as evidenced by the 6 false negatives. For "Upper Eyelid Obstructing," the classifier demonstrated high accuracy in detection and differentiation, with only 1 false negative and no false positives. In the "Image Too Dark" class, the model achieved a high true positive rate but exhibited misclassification issues, as seen in the 8 false positives. The "Dark Artifact" class showed the model's strong ability to identify dark artifacts and distinguish them from other conditions. Finally, in the "Image Not Centered" class, the classifier reasonably differentiated non-centered images from centered ones, but with room for improvement in reducing false negatives.

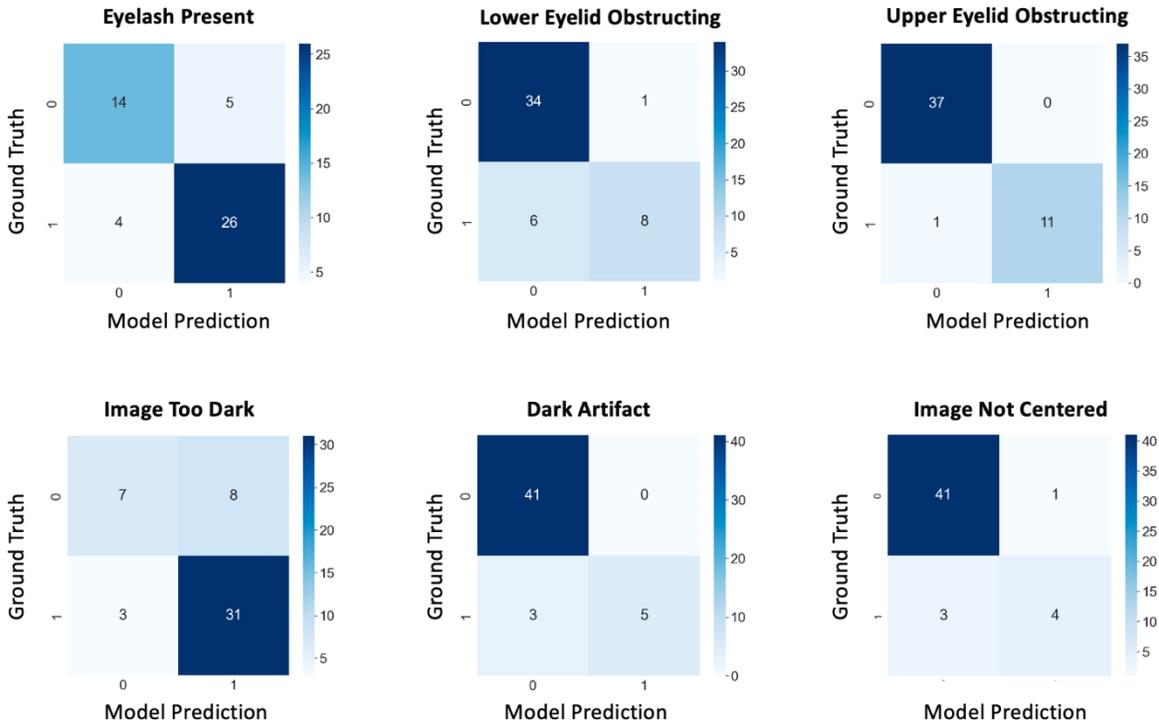

Figure 2: Confusion matrices for the prediction of the six labels.

Qualitative results are shown in Figure 3. For most input images, the model's predictions closely match the ground truth labels, emphasizing the classifier's effectiveness in recognizing and differentiating between distinct visual features present in the retina images, as shown in Figure 3. However, it is important to acknowledge that there still be some cases in which the classifier's performance falls short, as demonstrated in

Figure 4.

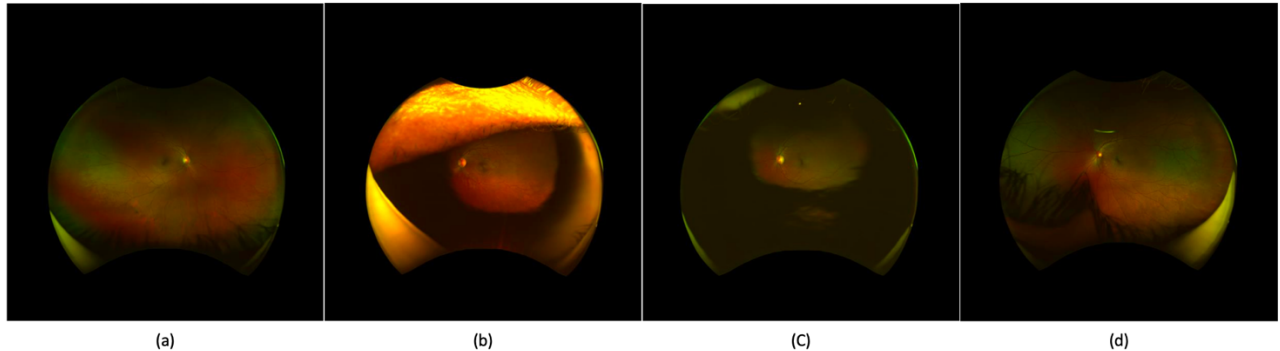

Figure 3: Visualization of the test images with accurate predictions. (a) Ground Truth Label: Eyelash Present, Image too dark; Model Prediction: Eyelash Present, Image too dark. (b) Ground Truth Label: Upper Eyelid obstructing, Image not centered; Model Prediction: Upper Eyelid obstructing, Image not centered. (c) Ground Truth Label: Dark Artifact; Model Prediction: Dark Artifact. (d) Ground Truth Label: Eyelash Present, Lower Eyelid obstructing, Image too dark; Model Prediction: Eyelash Present, Lower Eyelid obstructing, Image too dark.

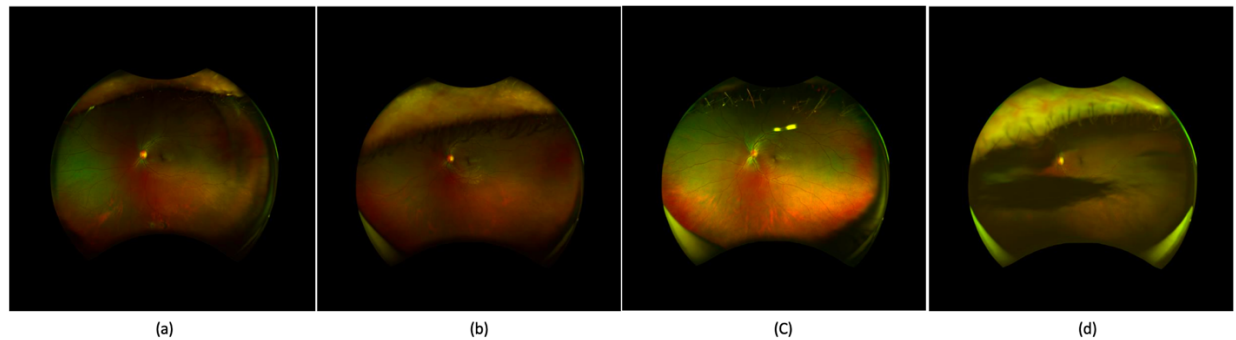

Figure 4: Visualization of the test images with wrong predictions. (a) Ground Truth Label: Eyelash Present, Image too dark; Model Prediction: Eyelash Present. (b) Ground Truth Label: Eyelash Present, Upper Eyelid Obstructing; Model Prediction: Upper Eyelid Obstructing. (c) Ground Truth Label: Eyelash Present, Image Not Centered; Model Prediction: Eyelash Present, Image Too Dark. (d) Ground Truth Label: Upper Eyelid Obstructing, Dark Artifact; Model Prediction: Eyelash Present, Upper Eyelid Obstructing.

In summary, the overall performance of our model, as evidenced by the close alignment between the ground truth labels and the model's predictions in the visualized retina images, is highly satisfactory. This suggests that our multi-label classification model is well-suited for the given domain and holds significant potential as a valuable tool for researchers and practitioners in the field.

# Discussion

In this study, we developed a multi-label classification model for the analysis of ocular images, addressing various image quality issues and obstructions. Our model's performance was assessed using multiple evaluation metrics, highlighting its strengths and limitations.

The model demonstrates high proficiency in detecting the presence of upper eyelid obstructions, dark artifacts, and off-centered images, while showing room for improvement in detecting cases with lower eyelid obstructions, eyelashes, and excessively dark images. The overall evaluation metrics, such as accuracy, precision, recall, and specificity, underscore the model's reliability and discriminative power.

The confusion matrices provide further insight into the model's performance across different class labels, identifying areas where misclassifications occur. By examining these results (Figure 4), we can identify the specific challenges the model faces in recognizing and differentiating between distinct visual features in ocular images. These failure cases can often be attributed to the qualitative and subjective nature of the expert image grades that may present a challenge for the classifier to accurately discern the correct labels. These insights can guide future research efforts to enhance the model's performance and address its limitations.

The qualitative results, consisting of visual comparisons between ground truth labels and model predictions, further emphasize the model's effectiveness in most cases. However, some failure cases reveal the need for further refinement and investigation. These failures might be attributed to the classifier's inability to adequately capture the complex image features, suggesting that a more sophisticated model architecture could potentially improve its performance. Moreover, the lack of data might limit the model's ability to generalize across a broad range of scenarios and conditions, indicating that expanding the dataset could improve the robustness and overall performance of the model.

As a result, future directions for this work could involve investigating other model architectures such an unsupervised methods such as a WGAN. These architectures have shown promise in capturing the complex features inherent in image data, and therefore may potentially enhance our model's performance. Furthermore, expanding our dataset with more diverse instances can bolster the model's generalization capabilities and improve its robustness to various scenarios. It would also be beneficial to validate our model's performance on different datasets, which would further demonstrate its versatility and applicability across diverse contexts. By implementing these strategies, we aim to refine and improve upon the current model, ultimately fostering more accurate and reliable predictions in the field of ocular image analysis.

This study focuses on automating artifact detection so that photographers of varying experience levels may receive real-time feedback on quality of image acquisition. There are several limitations to our study. First, the artifact labelling relied on creation and categorization of commonly encountered artifacts in UWF images by a single retina specialist expert grader (A.W.S.). Images and artifact adjudication may be affected by subjectivity of the single grader. Further, the automated artifact detection model in this study is exclusively trained on a cohort of SCD patients, and the algorithm may not be able to be extrapolated for use in cohorts of UWF-FP obtained in patients with other retinal diseases. For example, because sea fan neovascularization in PSR shows a predilection for the superior temporal retinal periphery,[16] images without ideal visualization of this region would be labelled as suboptimal, as compared to other retinal pathologies such as diabetic retinopathy in which the areas of interest in image capture may be more evenly distributed in the posterior pole and though the entire retinal periphery. Additionally, all images were obtained from a patient cohort in an internal tertiary care center on the Optos UWF Primary platform. Future studies will further validate the model on external data sets, with assignment of artifact "ground truth" labeling by multiple retina specialists to decrease subjectivity in grading.

## Conclusions

To our knowledge, this is the first study of its kind to evaluate automated artifact detection in UWF-FP in sickle cell retinopathy screening. This technique would limit the need for a patient to return for re-imaging, or for a patient's retinal images to be deemed ungradable due to suboptimal image capture, thus improving efficiency for both the photographer and the patient. Our technique shows potential to improve screening not only for PSR, but possibly for other retinal conditions by providing a photographer with real-time feedback as the presence of image artifact, and information as to what type of artifact is present. Refining sickle cell retinopathy screening methods through approaches such as automated artifact detection in UWF-FP is an important step in improving point-of-care access for patients with SCD and detection of sea-fan neovascularization, with potential to improve success and accuracy rates of SCR tele-screening programs in medically underserved regions worldwide with a high burden of SCD patients, which are more likely to have volunteers and/or novice photography personnel.

## Acknowledgements

This work was possible through a grant from the Johns Hopkins Malone Center for Engineering in Healthcare, the Retina Society, and the International Retina Research Foundation. The authors have no conflicts of interest.

# References

1. Kavanagh PL, Fasipe TA, Wun T. Sickle cell disease: A review. *JAMA*. 2022;328(1):57-68.

2. Downes SM, Hambleton IR, Chuang EL, Lois N, Serjeant GR, Bird AC. Incidence and natural history of proliferative sickle cell retinopathy: observations from a cohort study. *Ophthalmology*. 2005;112(11):1869-1875.

3. Chen RWS, Flynn HW Jr, Lee WH, et al. Vitreoretinal management and surgical outcomes in proliferative sickle retinopathy: a case series. *Am J Ophthalmol*. 2014;157(4):870-875.e1.

4. Farber MD, Jampol LM, Fox P, et al. A randomized clinical trial of scatter photocoagulation of proliferative sickle cell retinopathy. *Arch Ophthalmol*. 1991;109(3):363-367.

5. Yawn BP, Buchanan GR, Afenyi-Annan AN, et al. Management of sickle cell disease. *JAMA*. 2014;312(10):1033.

6. Cronin RM, Hankins JS, Byrd J, et al. Modifying factors of the health belief model associated with missed clinic appointments among individuals with sickle cell disease. *Hematology*. 2018;23(9):1-9.

7. Zulueta P, Minniti CP, Rai A, Toribio TJ, Moon JY, Mian UK. Routine ophthalmological examination rates in adults with sickle cell disease are low and must be improved. *Int J Environ Res Public Health*. 2023;20(4). doi:10.3390/ijerph20043451

8. Colombatti R, Montanaro M, Guasti F, et al. Comprehensive care for sickle cell disease immigrant patients: a reproducible model achieving high adherence to minimum standards of care. *Pediatr Blood Cancer*. 2012;59(7):1275-1279.

9. Ahmed I, Pradeep T, Goldberg MF, et al. Nonmydriatic ultra-Widefield fundus photography in a hematology clinic shows utility for screening of sickle cell retinopathy. *Am J Ophthalmol*. 2022;236:241-248.

10. Nagiel A, Lalane RA, Sadda SR, Schwartz SD. ULTRA-WIDEFIELD FUNDUS IMAGING: A Review of Clinical Applications and Future Trends. *Retina*. 2016;36(4):660-678.

11. Piel FB, Patil AP, Howes RE, et al. Global epidemiology of sickle haemoglobin in neonates: a contemporary geostatistical model-based map and population estimates. *Lancet*. 2013;381(9861):142-151.

12. He K, Zhang X, Ren S, Sun J. Deep residual learning for image recognition. In: *2016 IEEE Conference on Computer Vision and Pattern Recognition (CVPR)*. IEEE; 2016. doi:10.1109/cvpr.2016.90

13.	Ioffe S, Szegedy C. Batch Normalization: Accelerating deep network training by reducing internal covariate shift. *arXiv [csLG]*. Published online February 10, 2015. http://arxiv.org/abs/1502.03167

14.	Krizhevsky A, Sutskever I, Hinton GE. *ImageNet Classification with Deep Convolutional Neural Networks*. http://code.google.com/p/cuda-convnet/

15.	Deng J, Dong W, Socher R, Li LJ, Li K, Fei-Fei L. ImageNet: A large-scale hierarchical image database. In: *2009 IEEE Conference on Computer Vision and Pattern Recognition*. IEEE; 2009. doi:10.1109/cvpr.2009.5206848

16.	Goldberg MF. Classification and pathogenesis of proliferative sickle retinopathy. *Am J Ophthalmol*. 1971;71(3):649-665.